\documentclass{article}
\usepackage{spconf,amsmath,graphicx,amssymb,amsfonts}
\usepackage{algorithmicx}
\usepackage[ruled]{algorithm}
\usepackage{algpseudocode}
\usepackage{comment}
\usepackage{xcolor}
\usepackage{soul}
\usepackage{hyperref} 
\hypersetup{colorlinks,urlcolor=blue}

\definecolor{newcolor}{rgb}{.8,.349,.1}

\title{Learning diversified feature representations\\for facial expression recognition in the wild}

\name{Negar Heidari and Alexandros Iosifidis \thanks{This work received funding from the European Union’s Horizon 2020 research and innovation programme under grant agreement No. 871449 (OpenDR). This publication reflects the authors’ views only. The European Commission is not responsible for any use that may be made of the information it contains.}}
\address{Department of Electrical and Computer Engineering, Aarhus University, Denmark}

\begin{document}
\ninept
\maketitle
\begin{abstract}
Diversity of the features extracted by deep neural networks is important for enhancing the model generalization ability and accordingly its performance in different learning tasks.
Facial expression recognition in the wild has attracted interest in recent years due to the challenges existing in this area for extracting discriminative and informative features from occluded images in real-world scenarios. In this paper, we propose a mechanism to diversify the features extracted by CNN layers of state-of-the-art facial expression recognition architectures for enhancing the model capacity in learning discriminative features. To evaluate the effectiveness of the proposed approach, we incorporate this mechanism in two state-of-the-art models to (i) diversify local/global features in an attention-based model and (ii) diversify features extracted by different learners in an ensemble-based model. Experimental results on three well-known facial expression recognition in-the-wild datasets, AffectNet, FER+ and RAF-DB, show the effectiveness of our method, achieving state-of-the-art performance of $89.99\%$ on RAF-DB, $89.34\%$ on FER+ and the competitive accuracy of $60.02\%$ on AffectNet dataset. 

\end{abstract}

\section{Introduction}
\label{sec:intro}

Facial expression as a fundamental natural signal for human social communication plays an important role in different applications of artificial intelligence, such as Human Computer Interaction (HCI), healthcare, and driver fatigue monitoring. 
Deep Convolutional Neural Networks (CNNs) have led to considerable progress in automatic Facial Expression Recognition (FER) on large-scale datasets in real-world scenarios. 
FER methods aim to solve a visual perception problem by learning feature representations from facial images/videos to be classified as an emotional category, i.e. happiness, sadness, fear, anger, surprise, disgust, neutral, and contempt.  
In laboratory-controlled datasets, such as CK+ \cite{lucey2010extended} and JAFFE \cite{lyons1998coding}, where the facial images are in fixed frontal pose without any occlusion, FER methods have achieved excellent performance. However, these methods confront challenges for in-the-wild datasets, such as AffectNet \cite{mollahosseini2017affectnet}, FER+ \cite{barsoum2016training}, and RAF-DB \cite{li2017reliable}, where facial images come with illumination, occlusion and pose variations causing considerable change in facial appearance. 
To address that, many recent methods rely on transfer learning to exploit the feature representations learned for other visual perception tasks, such as object recognition, with well-designed networks, like ResNet-18 \cite{he2016deep}, trained on large datasets, like VGG-Face \cite{parkhi2015deep} and MS-Celeb-1M \cite{guo2016ms}, to be transferred for facial expression recognition in challenging in-the-wild datasets. 
However, considering that many face datasets are small and imbalanced, these deep neural networks are mostly over-parameterized and tend to overfit on the training data, which can degrade their generalization ability on unseen data. 

Increasing the diversity of features learned by different network layers/neurons has been recognized as an effective way to improve model generalization \cite{xie2017diverse}. It is theoretically shown in \cite{laakom2021within,laakom2021feature} that the within-layer activation diversity improves the generalization performance of neural networks and lowers the effect of overfitting. 
In this paper, we propose a mechanism for learning diversified facial feature representations by encouraging the learner to extract diverse spatial and channel-wise features. This mechanism can be used in different CNN architectures to increase the features diversity between layers or branches, spatial regions, and/or channels of feature maps. 
We incorporate our proposed optimization mechanism into two state-of-the-art models, i.e., the MA-Net \cite{zhao2021learning} and the ESR \cite{siqueira2020efficient}, and conduct experiments on three well-known in-the-wild datasets, i.e., AffectNet, FER+ and RAF-DB. Experimental results demonstrate the effectiveness of learning diversified features in improving the accuracy and generalization of the pretrained state-of-the-art models on new samples. 

The contributions of the paper can be summarized as follows: 
\begin{itemize}
    \item We propose a mechanism for learning diversified features in spatial and channel dimensions of CNNs to improve the model's accuracy in discriminating facial expressions. 
    \item We evaluate our feature extraction mechanism by incorporating it into two state-of-the-art models which have different properties, i.e., one benefits from a region-based attention mechanism and transfer learning, and the other one is an efficient ensemble-based architecture. In both cases, our diversified feature learning mechanisms boost the performance.
    \item Conducted experiments on three benchmark in-the-wild datasets, including the large-scale dataset AffectNet, indicate the effectiveness and adaptability of our method, which can be used in different types of models. Our code is publicly available at \href{https://github.com/negarhdr/Diversified-Facial-Expression-Recognition}{https://github.com/negarhdr/Diversified-Facial-Expression-Recognition}.
\end{itemize}

\section{Related works}\label{sec:related}\vspace{-0.2cm}
Recent studies are focused on addressing the challenges of in-the-wild facial expression recognition by 
training models with multi-pose examples \cite{zhang2018joint}, and extracting key facial features based on facial landmarks and region-based attention mechanisms \cite{li2018occlusion,wang2020region,heidari2021progressive}. 
Learning facial features from global and local perspectives simulates the human brain's perception mechanism and helps achieving better performance in visual perception problems. 
MA-Net \cite{zhao2021learning} is a global multi-scale and local attention network which extracts features with different receptive fields, to increase the diversity and robustness of global features.
This state-of-the-art method comprises of a backbone based on ResNet-18 for extracting preliminary features which are fed into a two-branch network with global multi-scale and local attention modules for high-level feature extraction. The first branch receives the preliminary feature maps as input and applies several multi-scale convolutions to extract both deeper semantic and shallower geometry features. The second branch of the network also receives the preliminary feature maps extracted by the backbone network as input, divides the feature maps into several local spatial regions without overlap, and then applies several parallel local attention networks to highlight the most important facial features in each region. At the end, a decision-level fusion strategy is employed to classify the extracted multi-scale and local attention features into different facial expression categories. 
However, this large network with 50.54 M parameters needs to be trained on a large dataset, and consistent with other state-of-the-art methods \cite{wang2020suppressing, wang2020region}, this network is first trained on MS-Celeb-1M dataset, and then finetuned on in-the-wild facial datasets AffectNet and RAF-DB. 

ESR \cite{siqueira2020efficient} has solved this issue by proposing an efficient ensemble-based method which reduces the residual generalization error on the AffectNet and FER+ datasets, and achieves state-of-the-art performance while training from scratch on these datasets. ESR model consists of two building blocks: 1) the base network which is composed of a stack of convolutional layers and is responsible for extracting low/middle-level features,  
2) the ensemble network composed of several network branches which are supposed to learn distinctive features. All branches in this ensemble module receive the same feature maps extracted by the base network as input, and they compete for a common resource which is the base network. 
The training algorithm of ESR starts with training the base network and one ensemble branch. Thereafter, more convolutional branches are added one by one while training, so that the base network leads and speeds up learning by providing all ensemble branches with shared preliminary feature maps which are suitable for all the branches. Therefore, this method reduces redundancy in low-level feature learning and focuses on learning high-level discriminative features to be classified. Finally, the input facial image is classified to an emotion category by fusing the predictions of all the ensemble branches and applying majority voting. 

In this paper we propose to complement discriminative feature extraction by increasing the features diversity between attention-regions, channel dimensions, and ensemble branches. In the next section, the diversified feature learning mechanism is introduced, and accordingly the modified learning mechanism for MA-Net and ESR methods is described.

\section{Proposed Method}\label{sec:proposed}\vspace{-0.2cm}
Diversity of feature representations is important in deep learning for enhancing the model generalization on unseen data, and improving model's accuracy in perceptual tasks by extracting non-redundant and discriminative features. 
Inspired by \cite{laakom2021within}, we propose to increase spatial and channel-wise feature diversity in CNN architectures and ensemble-based models for facial expression recognition. 
\begin{figure}[!t]
\centering
\centerline{\includegraphics[width=0.5\textwidth]{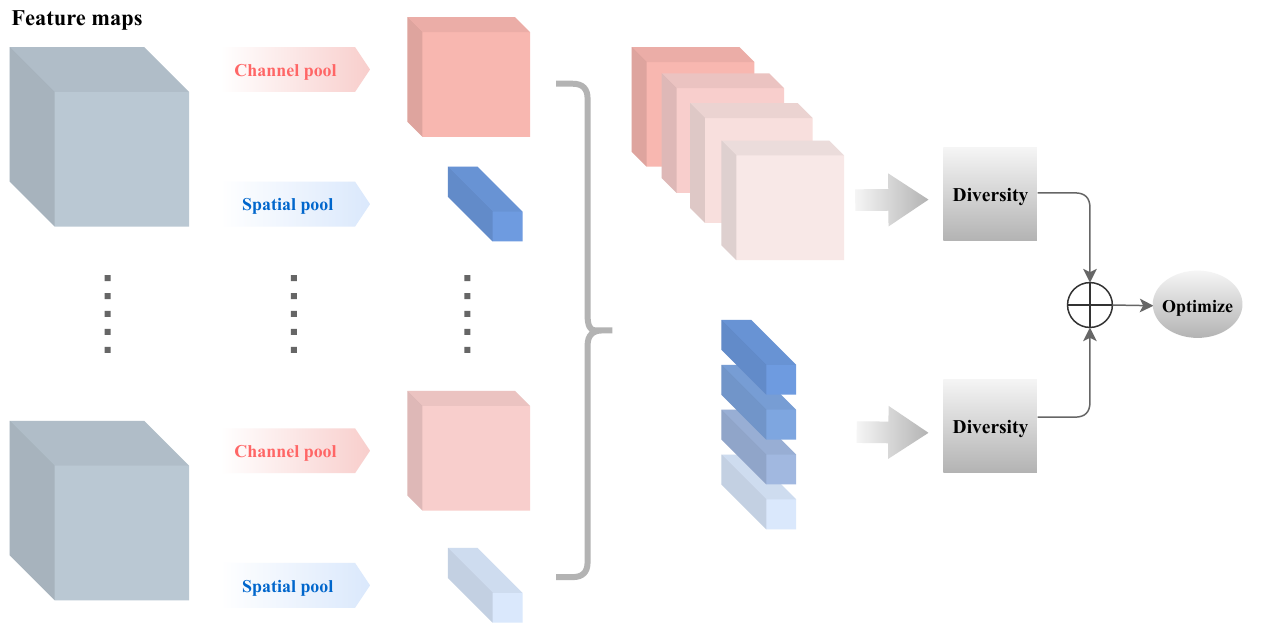}}
\caption{Illustration of diversity mechanism over the channel and spatial dimensions of feature maps. In this mechanism, spatial and channel pooling are applied on the feature maps of different learners/layers, and the diversity between different pooled features are then computed and optimized by the training algorithm.}
\label{fig:diversity}
\end{figure}

Let us assume that $\Phi_l, l \in \left \{ 1,2, ..., L \right \}$ is a feature map of size $C \times H \times W$ extracted by a CNN learner. The diversity between different feature maps obtained by different learners or different layers of a CNN model can be obtained in channel and spatial dimensions as illustrated in Figure \ref{fig:diversity} by first applying pooling on spatial and channel dimensions and then computing the average similarity between every two pooled feature maps $l$, $k$ using radial basis function as follows: 
\begin{equation}
\label{eq:sim}
    S_{lk} = \frac{1}{N} \sum_{i=1}^{N} exp(-\gamma \left \| \phi_l(\mathbf{x}_i) - \phi_k(\mathbf{x}_i) \right \|^{2}), 
\end{equation}
where $N$ denotes the number of samples from which feature maps are extracted, $\gamma$ is a hyperparameter, $\phi_l(\cdot)$ and $\phi_k(\cdot)$ denote the pooled feature maps of the $l^{th}$ and  $k^{th}$ learners, respectively. The feature maps are of size $1 \times H \times W$ and $C \times 1 \times 1$ when similarity is measured on spatial (Figure \ref{fig:diversity} top row) and channel (Figure \ref{fig:diversity} bottom row) dimensions, respectively.  
Similar feature maps indicate low diversity of the learner. Accordingly, using the pairwise similarities between feature maps, the model diversity is obtained by computing the determinant of the matrix $\mathbf{S}$ indicating pairwise similarities of learners as $S_{lk}$, i.e.,: 
\begin{equation}
\label{eq:div}
    \mathcal{D} = det(\mathbf{S}).
\end{equation}
The model can be optimized in an end-to-end manner by minimizing the combined loss function comprising of classification loss and diversity.  
That is, the overall loss value to be minimized is: 
\begin{equation}
    Loss = \mathcal{L} - (\mathcal{D}_{ch} + \mathcal{D}_{sp}), 
\end{equation}
where $\mathcal{L}$ denotes the cross-entropy classification loss, and $\mathcal{D}_{ch}$, $\mathcal{D}_{sp}$ denote the feature diversity computed through channel and spatial dimensions, respectively using Eq. (\ref{eq:sim}) and Eq. (\ref{eq:div}). 

\begin{figure}
\centering
\centerline{\includegraphics[width=0.5\textwidth]{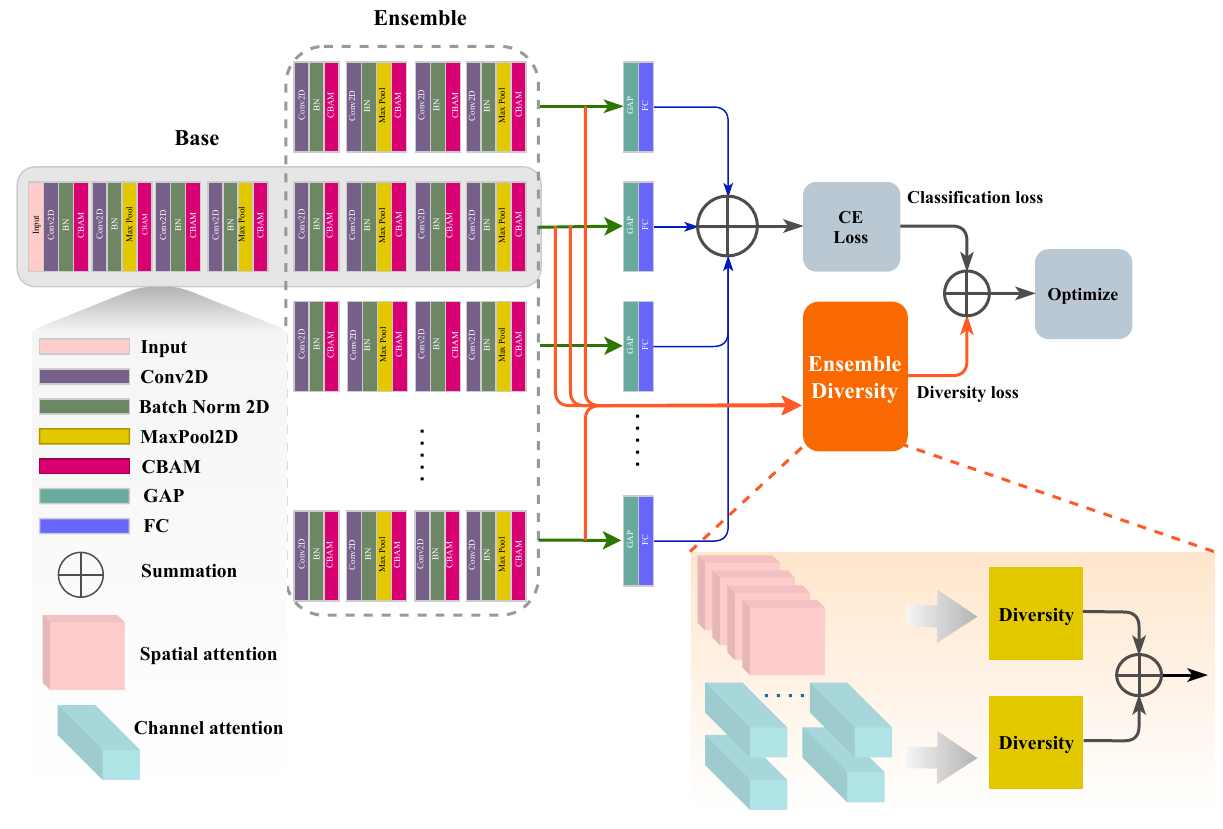}}
\caption{Illustration of the modified ESR structure with added CBAM attention modules in each layer and the ensemble diversity block which diversifies the channel and spatial attention maps of the ensemble branches.}
\label{fig:ESR}
\end{figure}

This mechanism can be used in CNN-based models to increase diversity between the feature maps at different levels. 
Considering the fact that diversity of learners is important in ensemble learning, encouraging each branch of ESR to learn complementary features of data can lead to better ensemble classification. To reach this goal, we modified the architecture of ESR by adding the CBAM attention mechanism \cite{woo2018cbam} into each layer of the network and maximizing the diversity of both channel and spatial attention maps between different branches. The combined loss function of the modified ESR model is defined as a summation of the diversity loss between branches and the cross-entropy loss of each branch as follows: 
\begin{equation}
    \mathcal{L}_{esr} = \sum_{b} \mathcal{L}(f_b(\mathbf{X})) - (\mathcal{D}_{ch} + \mathcal{D}_{sp}), 
\end{equation}
where $\mathcal{L}$ denotes the cross-entropy classification loss function for each of the ensemble learners $f_b$ which is combined with a negative summation of spatial and channel diversity of the whole ensemble model. In other words, by optimizing $\mathcal{L}_{esr}$, the features of ensemble branches (learner) are diversified, while each branch is encouraged to classify features with minimum loss. 
Figure \ref{fig:ESR} illustrates the new structure of ESR with added attention modules in each layer and our augmented module which computes the ensemble diversity to be optimized with cross entropy loss.

\begin{figure}
\centering
\centerline{\includegraphics[width=0.5\textwidth]{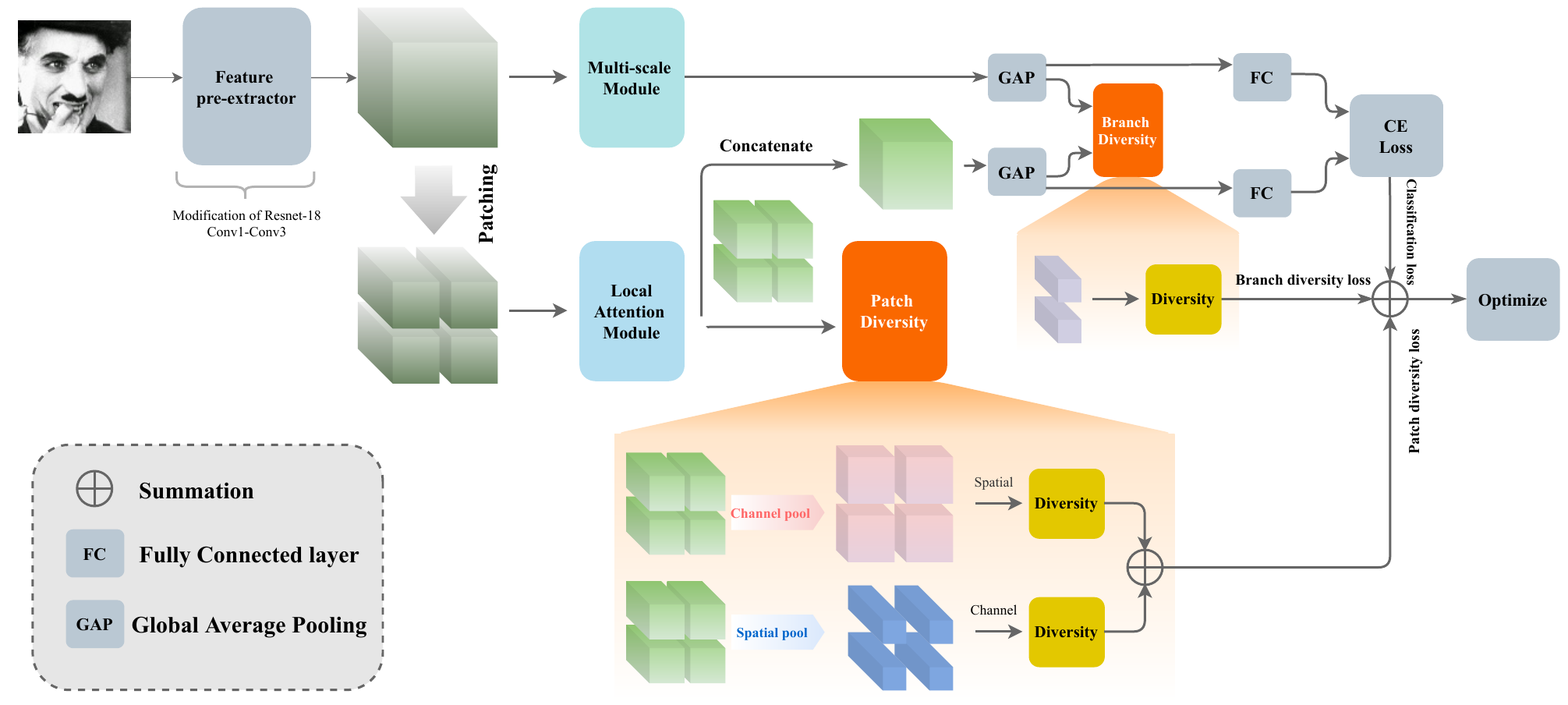}}
\caption{Illustration of the modified MA-Net structure by adding patch diversity and branch diversity blocks to diversify local region-based features in each feature map and also increase diversity of the extracted global and local features before passing them to the classification layers. }
\label{fig:MANeT}
\end{figure}

In MA-Net, the focus is on exploiting both local and global feature in two model branches. CBAM attention mechanism is originally employed in this method to highlight the key global and local facial regions for recognizing the expression. As illustrated in Figure \ref{fig:MANeT}, the first branch of the network employs the feature map tensor in its initial shape to extract the global features, but the second branch divides the feature map into four patches to learn and highlight the local feature in each of the patches separately. We modified MA-Net structure to encourage the local branch to learn diversified regional features. In this regard, channel and spatial pooling operations are applied on divided patches and the diversity between them are computed to be added to the model classification loss function. 
Besides, in order to make the two branches of the network as effective as ensemble learner, the global and the local features extracted by these two branches can be diversified as well. In this regard, the local feature patches are concatenated and introduced to the global average pooling layer, along with the global feature map, and the pooled features are diversified by the branch diversity block and then classified by the fully connected layer. The whole model is optimized by minimizing a combined loss function comprising of local and global classification loss in addition to the branch and patch diversity as follows: 
\begin{equation}
    Loss = \lambda \mathcal{L}_{local} + (1-\lambda) \mathcal{L}_{global} - (\mathcal{D}_{b} + (\mathcal{D}_{sp} + \mathcal{D}_{ch})), 
\end{equation}
where $\mathcal{L}_{local}$, $\mathcal{L}_{global}$ denote the cross-entropy classification loss in the local and global branches, respectively. $0 \le \lambda \le 1$ is a hyperparameter balancing the two parts and the best performance of MA-Net is obtained for $\lambda = 0.6$. $\mathcal{D}_b$ is the diversity between the two branches and $\mathcal{D}_{sp}$, $\mathcal{D}_{ch}$ indicate spatial and channel diversity between the local feature patches.

\section{Experiments}\label{sec:experiments}\vspace{-0.2cm}
We conducted experiments on three widely used in-the-wild datasets, AffectNet \cite{mollahosseini2017affectnet}, FER+ \cite{barsoum2016training}, RAF-DB \cite{li2017reliable}. 
\textbf{AffectNet} is the largest in-the-wild dataset containing more than one million images collected from the Internet by querying emotion keywords in different languages. Following the same experimental setting as in RAN \cite{wang2020region}, ESR \cite{siqueira2020efficient}, and SCN \cite{wang2020suppressing}, we used 450,000 images of this dataset which are manually annotated with 8 discrete expressions containing 6 basic ones (happiness, surprise, sadness, anger, disgust, fear) plus neutral and contempt. 287,568 images are used as training data and 4,000 images are used as test data. 
%
\textbf{FER+} is an extension of FER2013 \cite{goodfellow2013challenges} dataset, which is a large-scale dataset containing 35,887 facial images collected by Google search engine with 7 expressions. 
FER+ annotators re-labeled the FER2013 by crowd-sourcing and added contempt expression to the dataset. All the face images in this dataset are aligned and annotated with 8 expressions.  
\textbf{RAF-DB} dataset comprises of 30,000 facial images annotated with basic or compound expressions. Similar to the experimental setting in state-of-the-art methods, we used images with basic expressions, including 12,271 training and 3,068 test image, in our experiments. 

The experiments are conducted on PyTorch deep learning framework \cite{paszke2017automatic} with one GRX 1080-ti GPU, SGD optimizer with a momentum of 0.9 and cross-entropy loss function. We followed the experimental setting of ESR and MA-Net methods for reproducing their results and also training the modified version of the models on all the datasets. Table \ref{table:ablation} shows a summary of our experimental results for all the three datasets, and Tables \ref{table:AffectNet}, \ref{table:FER+}, \ref{table:RAF-DB} compare the performance of the state-of-the-art methods with our proposed diversified ESR and MA-Net models in AffectNet, FER+ and RAF-DB, respectively. 

\begin{table}[!h]
	\centering
	\caption{Comparisons of the classification accuracy of two state-of-the-art methods, ESR and MA-Net, and their modified versions with and without (channel and spatial) diversity computing on AffectNet, FER+, and RAF-DB datasets. $*$ indicates our proposed version of the method. \newline}  
	\label{table:ablation}
	\resizebox{\linewidth}{!}{
	\begin{tabular}{lcccccc}
		\hline
		Method & Attention & \multicolumn{2}{c}{Diversity} & \multicolumn{3}{c}{Dataset}\\
		
         &  & Spatial & Channel & AffectNet & FER+ & RAF-DB \\
        \hline
		ESR-9 \cite{siqueira2020efficient} & $\times$ & $\times$ & $\times$ & \textbf{59.3} & 87.17 & -\\ 
		ESR-9$^{*}$ & $\times$ & $\times$ & $\times$ & 58.8 & 88.40 & 77.96 \\
                    & $\checkmark$ & $\times$ & $\times$ & 58.95 & 88.56 & 82.39 \\
                    & $\checkmark$ & $\checkmark$ & $\times$ & 59.25 & 88.59 & 82.92 \\
                    & $\checkmark$ & $\checkmark$ & $\checkmark$ & \textbf{59.3} & \textbf{89.15} & \textbf{82.95} \\
        \hline
        ESR-15$^{*}$ & $\times$ & $\times$ & $\times$ & 58.7 & 88.59 & 77.5 \\
		             & $\checkmark$ & $\times$ & $\times$ & 59.25 & 88.78 & 82.82 \\
		             & $\checkmark$ & $\checkmark$ & $\times$ & 59.47 & 89.21 & 82.92 \\
		             & $\checkmark$ & $\checkmark$ & $\checkmark$ & \textbf{60.00} & \textbf{89.34} & \textbf{83.00} \\
		\hline
		MA-Net \cite{zhao2021learning} & $\checkmark$ & $\times$ & $\times$ & \textbf{60.29} & - & 88.4 \\
		MAN-Net$^{*}$ & $\checkmark$& $\times$ & $\times$ & 59.85 & 87.49 & 88.68 \\
	                  & $\checkmark$ & $\checkmark$ & $\checkmark$ & 60.02 & \textbf{88.34} & \textbf{89.99} \\
		
		\hline
	\end{tabular}}
\end{table}

\begin{table}[!t]
	\centering
	\caption{Comparisons of the classification accuracy of the state-of-the-arts with our proposed version of ESR and MA-Net methods on AffectNet dataset with 8 classes. $*$ indicates our proposed version of the method. \newline} 
	\label{table:AffectNet}
	\resizebox{0.67\linewidth}{!}{
	\begin{tabular}{lcc}
		\cline{1-3}
		Methods  & Pretrained & Acc.(\%) \\
		\hline
		
		MobileNet \cite{hewitt2018cnn} & - & 56.00 \\
		VGGNet \cite{hewitt2018cnn} & - & 58.00 \\
		AlexNet-WL \cite{mollahosseini2017affectnet} & - & 58.00 \\
		RAN \cite{wang2020region} & MS-Celeb-1M & 59.50 \\
		SCN \cite{wang2020suppressing} & MS-Celeb-1M & 60.23 \\
		
		\hline \hline
		ESR-9 \cite{siqueira2020efficient} & AffectNet & 59.30 \\ 
		ESR-9$^{*}$ & AffectNet & 59.30 \\
		ESR-15$^{*}$ & AffectNet & \textbf{60.00} \\
		
		\hline
		MA-Net \cite{zhao2021learning} & MS-Celeb-1M & \textbf{60.29} \\
	    MA-Net$^{*}$ & MS-Celeb-1M & 60.02 \\ 
		\hline
	\end{tabular}}
\end{table}

\begin{table}[!t]
	\centering
	\caption{Comparisons of the classification accuracy of the state-of-the-arts with our proposed version of ESR and MA-Net methods on FER+ dataset with 8 classes. $*$ indicates our proposed version of the method. \newline}  
	\label{table:FER+}
	\resizebox{0.67\linewidth}{!}{
	\begin{tabular}{lcc}
		\hline
		\cline{1-3}
		Methods & Pretrained & Acc.(\%) \\
		\hline
        TFE-JL \cite{li2018facial} & - & 84.30 \\ 
        PLD \cite{barsoum2016training} & - & 85.10 \\ 
        SHCNN \cite{miao2019recognizing} & - & 86.54 \\
        SeNet50 \cite{albanie2018emotion} & VGG-Face2 \cite{cao2018vggface2} & 88.80 \\ 
        RAN \cite{wang2020region} & MS-Celeb-1M  & 88.55 \\
                                  & VGG-Face \cite{parkhi2015deep} & 89.16 \\
        SCN \cite{wang2020suppressing} & MS-Celeb-1M  & 88.01 \\ 
        
		\hline \hline
		ESR-9 \cite{siqueira2020efficient} & AffectNet & 87.17 \\
        ESR-9$^{*}$ & AffectNet & \textbf{89.15} \\
		ESR-15$^{*}$ & AffectNet & \textbf{89.34} \\
		
		\hline
		MAN-Net \cite{zhao2021learning} & MS-Celeb-1M & - \\
	    MAN-Net$^{*}$ & MS-Celeb-1M & 88.34 \\
		
		\hline
	\end{tabular}}
\end{table}

\begin{table}[!t]
	\centering
    	\caption{Comparisons of the classification accuracy of the state-of-the-arts with our proposed version of ESR and MA-Net methods on RAF-DB dataset with 7 classes. $*$ indicates our proposed version of the method. \newline} 
    	\label{table:RAF-DB}
    	 \resizebox{0.67\linewidth}{!}{
        	\begin{tabular}{lcc}
    	        \cline{1-3}
        		Methods  & Pretrained & Acc.(\%) \\
        		\hline
        		DLP-CNN \cite{li2017reliable} & - & 84.22 \\ 
        		IPA2LT \cite{zeng2018facial} & AffectNet &  86.77 \\ %
        		gACNN \cite{li2018occlusion} & AffectNet & 85.07 \\ 
        		LDL-ALSG \cite{chen2020label} & AffectNet & 85.53 \\ 
        		RAN \cite{wang2020region} & MS-Celeb-1M & 86.90 \\ 
        		SCN \cite{wang2020suppressing} & MS-Celeb-1M  & 87.03 \\
        		
        		\hline \hline
        		ESR-9 \cite{siqueira2020efficient} & AffectNet & - \\
        	    ESR-9$^{*}$ & AffectNet & 82.95 \\
        	    ESR-15$^{*}$ & AffectNet & 83.00\\
        	    
        	    \hline
        		MA-Net \cite{zhao2021learning} & MS-Celeb-1M & 88.40 \\ 
        		MA-Net$^{*}$  & MS-Celeb-1M & \textbf{89.99}\\ 
    		    \hline
    	    \end{tabular}}
\end{table}

To reproduce the ESR results with 9 ensemble branches, referred as ESR-9, the model is trained from scratch on AffectNet dataset and then finetuned on FER+. This led to $58.8\%$ accuracy on AffectNet which is $0.5\%$ less than the reported result in \cite{siqueira2020efficient}, however our reproduced result for FER+ is around $0.85\%$ higher than their originally reported accuracy. 
Although the performance of this method is not reported in \cite{siqueira2020efficient} for RAF-DB dataset, we finetuned the pretrained ESR-9 model on RAF-DB as well. 
To evaluate the effect of attention layers added to the ESR structure, we did an ablation study to compare the performance of ESR-9 with and without CBAM attention layers, and spatial/channel-wise diversities. According to the results reported in Table \ref{table:ablation}, the best ensemble classification accuracy is obtained by adding CBAM attention layers, as well as maximizing the diversity of both spatial and channel-wise  attention between ensemble branches. 

It is mentioned in \cite{siqueira2020efficient} that adding more than $9$ branches to ESR does not improve the performance. However, we assume that increasing the feature diversity between branches increases the model capacity for learning features with more ensemble branches. In this regard, we increased the number of branches both in the original ESR architecture and in our proposed version of ESR and trained the ESR-15 models on AffectNet and finetuned them on FER+ and RAF-DB. The results in Table \ref{table:ablation} confirm our assumption, and indicate the improved performance of ESR-15 for all the three datasets compared to ESR-9. 
It should be noted that the maximum number of branches is chosen empirically and in some cases, the best performance is achieved in earlier branches so that at inference time, we can get the results with early exits.
Considering the results in Table \ref{table:FER+}, ESR-15$^{*}$, which is our modified version of ESR-15 with diversified features, outperforms all the state-of-the-arts with no need to be pretrained on large-scale datasets like MS-Celeb-1M or VGG-Face. 

Since MA-Net original structure includes CBAM attention layers, we did not modify the structure of the layers in this model. MA-Net is first trained on MS-Celeb-1M dataset, and the pretrained weights are then finetuned on AffectNet and RAF-DB datasets. In our experiments, we used the available pretrained weights of MA-Net on MS-Celeb-1M, provided by the authors, and finetuned the weights on all the three datasets. However the reproduced classification accuracy on AffectNet is $0.44\%$ less than their reported accuracy of $60.29\%$, while for RAF-DB dataset we get $0.28\%$ higher accuracy than the original one. After augmenting branch and patch diversity blocks into the MA-Net structure and diversifying local and global features, we achieved the state-of-the-art performance of $89.99\%$ on RAF-DB dataset which outperforms all the other state-of-the-arts listed in Table \ref{table:RAF-DB}.

\section{Conclusion}\label{sec:conclusion}\vspace{-0.3cm}
In this paper we proposed a mechanism to diversify features extracted by different CNN learners for facial expression recognition. We targeted two state-of-the-art methods based on ensemble learning and multi-scale attention networks to evaluate the effect of learning diversified features in performance. Experimental results show that diversifying features extracted by different ensemble learners can enhance the overall ensemble classification performance while increasing the model capacity to include more learners for feature extraction. Furthermore, diversifying local regional features extracted by a CNN learner improves the model performance in exploiting local features and classifying facial images. 

\bibliographystyle{IEEEbib}
\bibliography{root}

\end{document}